\documentclass[conference]{IEEEtran}
\IEEEoverridecommandlockouts
\usepackage{cite}
\usepackage{amsmath,amssymb,amsfonts}
\usepackage{algorithmic}
\usepackage{graphicx}
\usepackage{textcomp}
\usepackage{xcolor}
\usepackage{pifont}
\usepackage{float}
\usepackage{url}
\usepackage{graphicx}
\usepackage{subcaption}
\usepackage{tabularx} % add this in your preamble

\usepackage[utf8]{inputenc}
\usepackage{newunicodechar}
\newunicodechar{≈}{\approx}

\usepackage{hyperref}
\hypersetup{
    colorlinks=true,
    allcolors=blue
}

\usepackage{afterpage}
\usepackage{booktabs}
\usepackage{tabularray}
\usepackage{microtype}
\usepackage{multirow}
\usepackage{array}
\newcommand{\PreserveBackslash}[1]{\let\temp=\\#1\let\\=\temp}
\newcolumntype{C}[1]{>{\PreserveBackslash\centering}p{#1}}
\newcolumntype{R}[1]{>{\PreserveBackslash\raggedleft}p{#1}}
\newcolumntype{L}[1]{>{\PreserveBackslash\raggedright}p{#1}}
\usepackage{booktabs}
\def\BibTeX{{\rm B\kern-.05em{\sc i\kern-.025em b}\kern-.08em
    T\kern-.1667em\lower.7ex\hbox{E}\kern-.125emX}}
\begin{document}

\title{Bi-cLSTM: Residual-Corrected Bidirectional LSTM for Aero-Engine RUL Estimation}

%\author{(Anonymous \confName  submission)}

%\author{ 
%    \IEEEauthorblockN{
%    Faria Ahmed,
%    Rafi Hassan Chowdhury, 
%    Sabbir Ahmed\\}

%    \IEEEauthorblockA{
%        Department of Computer Science and Engineering, Islamic University of Technology, Gazipur 1704, Bangladesh}

%    \IEEEauthorblockA{\{fariaahmed, rafihassan, sabbirahmed\}@iut-dhaka.edu
%    } 
%}
%\author{Anonymous ICCIT Submission}
\author{
    \IEEEauthorblockN{
        Rafi Hassan Chowdhury\textsuperscript{1},
        Nabil Daiyan\textsuperscript{3},
        Faria Ahmed\textsuperscript{1},
        Md Redwan Iqbal\textsuperscript{4},
        Morsalin Sheikh\textsuperscript{4}
    }

    \IEEEauthorblockA{
        \textsuperscript{2}Department of Computer Science and Engineering, 
        Islamic University of Technology, Gazipur 1704, Bangladesh\\
        \textit{\{rafihassan, fariaahmed\}@iut-dhaka.edu}
    }

    \IEEEauthorblockA{
        \textsuperscript{3}Department of Computer Science and Engineering, 
        Rajshahi University of Engineering and Technology, Rajshahi, Bangladesh\\
        \textit{1803014@student.ruet.ac.bd}
    }

    \IEEEauthorblockA{
        \textsuperscript{4}Department of Aeronautical Engineering, 
        Military Institute of Science and Technology, Dhaka 1216, Bangladesh\\
        \textit{\{redwan400900, morsalin.ae.mist\}@gmail.com}
    }
}

%%%%%%% ADDING CONFERENCE info in TOP-LEFT Corner %%%%%%%%%%
%  Source: https://tex.stackexchange.com/questions/561793/how-to-get-ieee-conference-template-to-show-conference-name-in-header

%\makeatletter
%\let\old@ps@IEEEtitlepagestyle\ps@IEEEtitlepagestyle
%\def\confheader#1{%
%    % for the first page
%    \def\ps@IEEEtitlepagestyle{%
%        \old@ps@IEEEtitlepagestyle%
%        \def\@oddhead{\strut\hfill#1\hfill\strut}%
%        \def\@evenhead{\strut\hfill#1\hfill\strut}%
%    }%
%    \ps@headings%
%}

\makeatletter
\let\old@ps@IEEEtitlepagestyle\ps@IEEEtitlepagestyle
\def\confheader#1{%
    % for the first page
    \def\ps@IEEEtitlepagestyle{%
        \old@ps@IEEEtitlepagestyle%
        \def\@oddhead{\strut\hfill#1\hfill\strut}%
        \def\@evenhead{\strut\hfill#1\hfill\strut}%
    }%
    \ps@headings%
}

\makeatother

\confheader{
        \parbox{20cm}{2025 28th International Conference on Computer and Information Technology (ICCIT)\\
        19-21 December 2025, Cox’s Bazar, Bangladesh}
}

\IEEEpubid{
\begin{minipage}[t]{\textwidth}\ \\[10pt]
       \small{979-8-3315-7867-1/25/\$31.00 \copyright2025 IEEE }
\end{minipage}
}

\makeatother

%\confheader{
%        \parbox{20cm}{2024 27th International Conference on Computer and Information Technology (ICCIT)\\
%$$        20-22 December 2024, Cox’s Bazar, Bangladesh 
%}
%}

% \IEEEpubid{
% \begin{minipage}[t]{\textwidth}\ \\[10pt]
%       \small{979-8-3315-1909-4/24/\$31.00 \copyright2024 IEEE }
% \end{minipage}
% }

%%%%%%%%% FOR ARXIV %%%%%%%%%%%

%\IEEEpubid{
%\begin{minipage}[t]{\textwidth}\ \\%[10pt]
%      \small{Accepted in 27th ICCIT \copyright2024 IEEE }  
%\end{minipage}
%}

\maketitle
\begin{abstract}
Accurate Remaining Useful Life (RUL) prediction is a key requirement for effective Prognostics and Health Management (PHM) in safety-critical systems such as aero-engines. Existing deep learning approaches, particularly LSTM-based models, often struggle to generalize across varying operating conditions and are sensitive to noise in multivariate sensor data. To address these challenges, we propose a novel \textbf{Bidirectional Residual Corrected LSTM (Bi-cLSTM)} model for robust RUL estimation. The proposed architecture combines bidirectional temporal modeling with an adaptive residual correction mechanism to iteratively refine sequence representations. In addition, we introduce a condition-aware preprocessing pipeline incorporating regime-based normalization, feature selection, and exponential smoothing to improve robustness under complex operating environments. Extensive experiments on all four subsets of the NASA C-MAPSS dataset demonstrate that the proposed Bi-cLSTM consistently outperforms LSTM-based baselines and achieves competitive state-of-the-art performance, particularly on challenging multi-condition scenarios. These results highlight the effectiveness of combining bidirectional temporal learning with residual correction for reliable RUL prediction.
\end{abstract}


\section{Introduction}

The commercial aerospace sector has a global industry of \$ 1.5 trillion. 40\% of airline maintenance expenses are spent on engine maintenance \cite{pwc2025aviation}. Although aircraft engines rarely shut down, the operational and financial consequences are very high. Accurately estimating the RUL earlier helps operators act proactively rather than reflectively. In this way, we can reduce unplanned downtime and operational costs.

Recent advances in deep learning have substantially improved the prediction of Remaining Useful Life (RUL). Asif et al. \cite{Asif2022IEEEAccess} proposed a deep learning model using the C-MAPSS dataset. This work is focused on turbofan engine RUL prediction and achieves accuracy improvement through effective spatiotemporal feature extraction and computational efficiency. Zhang et al. \cite{zhang2023multiscale} introduced a deep equilibrium model with iterative fixed-point computation. It shows better results in estimating complex temporal degradation patterns, but higher computational demands. Liu et al. presented an optimized predictive framework enhanced accuracy but lacked explicit techniques for noise and redundancy mitigation \cite{liu2024multi}. Hong et al. combined explainability with dimensionality reduction to increase model interpretability, though this sometimes sacrificed information and increased computational overhead \cite{s20226626}.

Despite considerable progress, many existing RUL models based on classical machine learning and deep neural networks, including LSTM and hybrid models, face challenges such as overfitting, sensitivity to noise \cite{Asif2022IEEEAccess}, and limited generalization to varied operating conditions \cite{zhang2023multiscale, liu2024multi}.

We introduce an enhanced Bidirectional Corrected LSTM (Bi\_cLSTM) for robust Remaining Useful Life (RUL) prediction. The model integrates bidirectional temporal modeling with a residual correction mechanism to refine predictions at each step. A preprocessing pipeline including condition aware normalization, Random Forest feature selection, and exponential smoothing reduces noise and redundancy. On the C-MAPSS dataset, Bi-cLSTM surpasses LSTM, cLSTM, and Bi-LSTM baselines, achieving lower RMSE and higher R², particularly on the challenging FD002 and FD004 subsets, demonstrating its ability to capture long-range temporal and contextual dependencies in complex operating conditions. \cite{gan2024adaptive, 11013930, liu2024enhancing, 11022513, mo2023evolutionary, 11022615, chowdhury2024mangoleafvit, ZHANG2024109662, wang2024aeroengine}

The remainder of this paper reviews related work, discusses a detailed proposed methodology, presents comprehensive experimental results and comparative analyses, and concludes with key findings, limitations, and future research directions.

\section{Methodology} \label{Methodology}

\subsection{Dataset}

We experimented on the Commercial Modular Aero-Propulsion System Simulation (C-MAPSS) dataset \cite{Saxena2008}, a widely used benchmark for Remaining Useful Life (RUL) prediction, provided by NASA's Prognostics Center of Excellence. The dataset simulates turbofan engine degradation under various operating conditions and fault modes, with 21 sensor signals (e.g., temperatures, pressures, rotational speeds) and three different operational settings. The task is to predict RUL based on these sensor measurements.

The dataset is of four subsets: FD001, FD002, FD003, and FD004, each have different levels of complexity. FD001 and FD003 have simpler configurations with one operating condition, on the other hand FD002 and FD004 include multiple conditions and fault modes. The number of engines and testing trajectories vary across subsets, as detailed in Table \ref{tab:dataset_subunits}. Sensor signals are normalized, and the maximum RUL is capped at 125 cycles to avoid bias from early stable phases.

Output parameters, such as fan speed, pressure, and temperature are summarized in Table \ref{tab:output_parameters}. To ensure robust evaluation, the training and validation sets for each subset are split without any overlap with the test set, following established protocols to avoid data leakage.

\begin{table}[t]
\centering
\caption{Each Sub-Unit of the C-MAPSS Dataset}
\label{tab:dataset_subunits}
\scriptsize
\begin{tabular}{l c c c c}
\toprule
\textbf{Dataset} & \textbf{FD001} & \textbf{FD002} & \textbf{FD003} & \textbf{FD004} \\
\midrule
Engines in training set & 100 & 260 & 100 & 249 \\
Engines in testing set & 100 & 259 & 100 & 248 \\
Training trajectories & 17,731 & 48,558 & 21,120 & 56,815 \\
Testing trajectories & 100 & 259 & 100 & 248 \\
Max/min cycle for train & 362/128 & 378/128 & 525/145 & 543/128 \\
Max/min cycle for test & 303/31 & 367/21 & 475/38 & 486/19 \\
Operating Conditions & 1 & 6 & 1 & 6 \\
Fault Modes & 1 & 1 & 2 & 2 \\
\bottomrule
\end{tabular}
\end{table}

\subsection{Dataset Preprocessing}

We utilized the NASA C-MAPSS turbofan engine degradation dataset, which provides multivariate time-series sensor measurements from engines operated until failure. Each record includes an engine identifier, a cycle index, three operational settings, and twenty-one sensor channels. The Remaining Useful Life (RUL) was defined as the number of cycles left until failure, computed as

\begin{equation}
    RUL_{i,t} = \max \left(0, \, T_i - t \right),
\end{equation}

where $T_i$ is the final cycle of engine $i$. For the test set, the provided offsets were incorporated into the terminal RUL values. To reduce extreme variations and improve model stability, all RUL values were clipped at 125 cycles following the standard practice in prognostics studies \cite{Zheng2017}.

Normalization was performed to account for distributional differences across operating conditions. For single-condition subsets (FD001 and FD003), a global z-score standardization was applied to all sensor channels:

\begin{equation}
    z = \frac{x - \mu}{\sigma},
\end{equation}

\begin{table}[t]
\centering
\caption{Output Parameters of the C-MAPSS Turbofan Engine Datasets}
\label{tab:output_parameters}
\scriptsize
\begin{tabular}{l l}
\toprule
\textbf{Sensor Parameter} & \textbf{Description with Units} \\
\midrule
T2 & Total Temperature in fan inlet (°R) \\
T24 & Total Temperature at LPC outlet (°R) \\
T30 & Total Temperature at HPC outlet (°R) \\
T50 & Total Temperature at LPT outlet (°R) \\
P2 & Pressure at fan inlet (psia) \\
P15 & Total pressure in bypass-duct (psia) \\
P30 & Total pressure at HPC outlet (psia) \\
Nf & Physical fan speed (rpm) \\
Nc & Physical core speed (rpm) \\
Epr & Engine pressure ratio (-) \\
Ps30 & Static pressure at HPC outlet (psia) \\
Phi & Ratio of fuel flow to Ps30 (psi) \\
NRf & Corrected fan speed (rpm) \\
Nrc & Corrected core speed (rpm) \\
BPR & Bypass ratio (-) \\
farB & Burner fuel air ratio (-) \\
htBleed & Bleed enthalpy (-) \\
NF-dmd & Demanded fan speed (rpm) \\
PCNR-dmd & Demanded corrected fan speed (rpm) \\
W31 & HPT coolant bleed (lbm/s) \\
W32 & LPT coolant bleed (lbm/s) \\
\bottomrule
\end{tabular}
\end{table}

where $\mu$ and $\sigma$ are computed from the training set. For multi-condition subsets (FD002 and FD004), the three operational settings ($os1$, $os2$, $os3$) were clustered into six regimes using K-means, and regime-specific normalization was applied within each cluster. The operational settings themselves were globally standardized to ensure consistency across engines.

To reduce redundancy, a Random Forest Regressor was used for feature selection. Sensors with importance scores below $10^{-3}$ were removed, leaving only the most informative variables. In addition, sensor measurements were smoothed using an exponentially weighted moving average (EWMA):

\begin{equation}
    \hat{x}_t = \beta \hat{x}_{t-1} + (1 - \beta)x_t,
\end{equation}

with a smoothing factor $\beta = 0.98$. Both raw and smoothed features were retained to capture short-term fluctuations as well as long-term degradation trends.

Sequential inputs were constructed using a sliding-window approach. Each subsequence comprised $W=15$ consecutive cycles, and the label was defined as the RUL at the final cycle of the window. Early cycles with $t \leq 10$ were excluded to avoid unstable initialization effects. For evaluation, only the final subsequence from each test unit was retained. The dataset was split  into 80\% training and 20\% validation sets. The final input tensors were structured as

\begin{equation}
    \mathbf{X} \in \mathbb{R}^{B \times W \times F},
\end{equation}

where $B$ is the batch size, $W=15$ is the sequence length, and $F$ is the number of selected features.

\subsection{Corrector Module}

The \textbf{Corrector} module is a residual correction mechanism designed to refine the outputs of sequential models, such as LSTMs and Bi-LSTMs, by learning an adaptive correction vector at each time step. This correction vector \( \mathbf{C}_t \) is added to the model's hidden state \( \mathbf{h}_t \) to improve prediction accuracy, particularly in tasks with noisy or incomplete data. The corrected output is given by:

\begin{equation}
    \mathbf{h}_t^{\text{corrected}} = \text{LayerNorm}(\mathbf{h}_t + \mathbf{C}_t)
\end{equation}

where \( \mathbf{C}_t = f(\mathbf{h}_t, \mathbf{x}_t) \) is the correction term computed by a small feedforward network, and \( \text{LayerNorm} \) ensures stability by normalizing the corrected output.

During training, the Corrector is integrated into the model, and its parameters are learned through backpropagation using the same loss function as the overall network. The model minimizes the Mean Squared Error (MSE) between the corrected predictions and the true labels. The inclusion of the Corrector improves temporal feature refinement, mitigates the impact of noise, and stabilizes the training process, making it particularly effective for predictive tasks like Remaining Useful Life (RUL) estimation.

\begin{figure*}[!t]
    \centering
    \includegraphics[width=\linewidth]{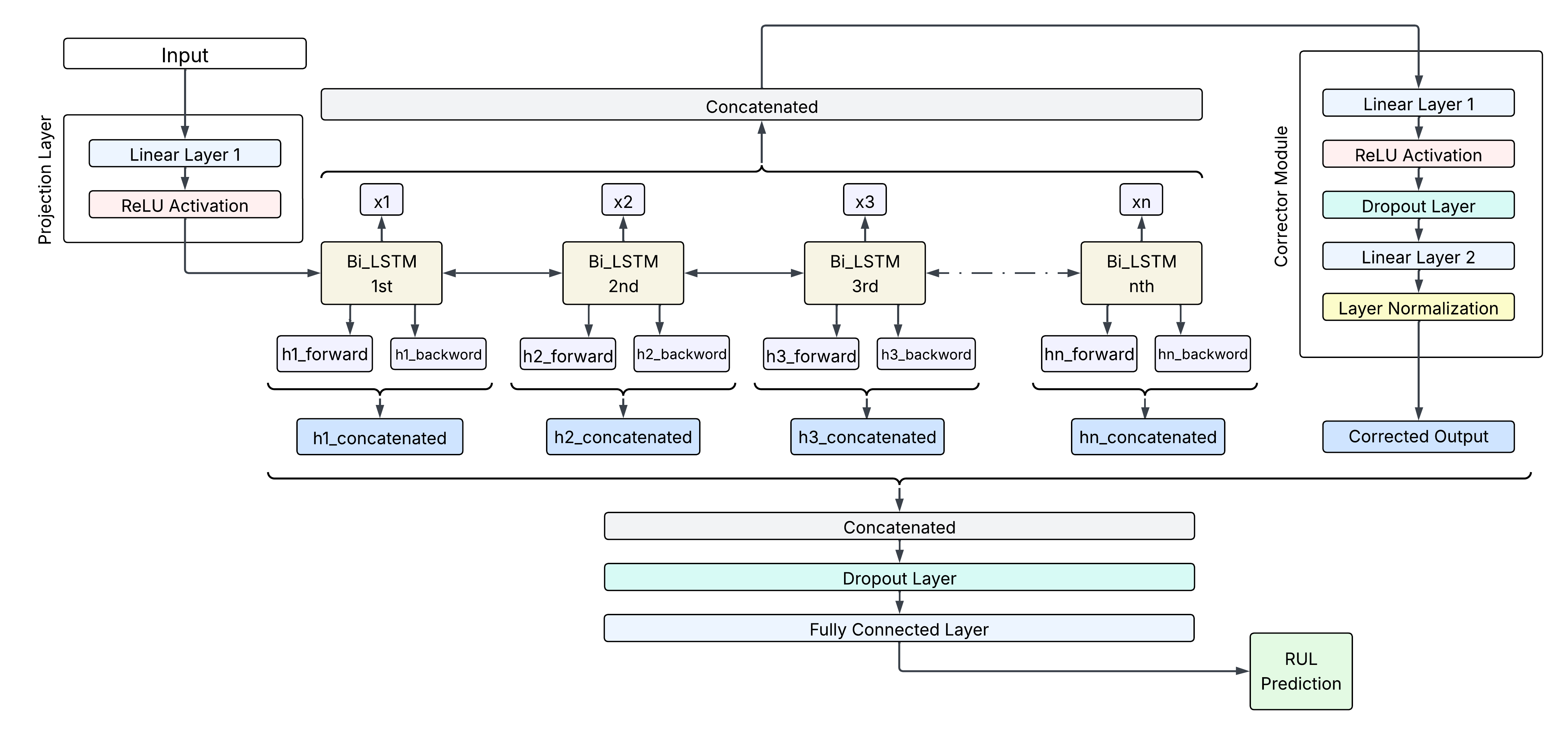}
    \caption{Proposed Bi-cLSTM framework with a residual Corrector module. Input features are projected, processed by bidirectional LSTMs, refined through adaptive correction, and mapped to Remaining Useful Life (RUL) predictions.}
    \label{fig:overview_of_proposed_framework}
\end{figure*}

\subsection{Proposed Bi-cLSTM (Bidirectional Corrected LSTM)}

The Bi-cLSTM model integrates bidirectional LSTM networks with a residual correction mechanism to enhance the prediction accuracy of sequential tasks, such as Remaining Useful Life (RUL) estimation. The primary goal of the architecture is to capture complex temporal dependencies in sequential data, while the residual correction mechanism refines predictions at each time step, compensating for potential inaccuracies in the LSTM's raw output.

The model starts by passing the input features \( \mathbf{x}_t \in \mathbb{R}^d \) through a projection layer that maps the raw input into a higher-dimensional space. This transformation enables the model to learn richer representations of the input sequence, which allows it to better capture complex temporal relationships inherent in the data. The projection is defined as:

\[
\mathbf{z}_t = \text{ReLU}(\mathbf{W}_p \mathbf{x}_t + \mathbf{b}_p)
\]

where \( \mathbf{W}_p \) and \( \mathbf{b}_p \) are the projection matrix and bias term, respectively, and ReLU is a non-linear activation function that introduces non-linearity into the transformation. This non-linearity enables the model to better handle complex data patterns that are not linearly separable.

Following the projection, the sequence is processed by multiple bidirectional LSTM blocks. The bidirectional LSTM captures both past and future dependencies by processing the input sequence in both forward and backward directions. This bidirectional processing is essential for tasks where both historical and future context are equally important for accurate prediction. Each LSTM block learns the temporal dependencies at each time step, producing hidden states \( \mathbf{h}_t \) and cell states \( \mathbf{c}_t \) as follows:

\[
\mathbf{h}_t, \mathbf{c}_t = \text{LSTM}(\mathbf{z}_t, \mathbf{h}_{t-1}, \mathbf{c}_{t-1})
\]

After processing through the LSTM blocks, the output is passed to the Corrector module, which computes a correction term \( \mathbf{C}_t \) for each time step. This correction is added to the LSTM output to refine the predictions. The final output from each block is then concatenated, and the RUL prediction is made through a fully connected layer:

\[
\hat{y}_t = \mathbf{W}_h \cdot \mathbf{h}_t^{\text{corrected}} + \mathbf{b}_h
\]

The model is trained end-to-end using a Mean Squared Error (MSE) loss function, minimizing the difference between the predicted and true RUL values. Training is performed using backpropagation through time (BPTT) with the Adam optimizer, and early stopping is applied to prevent overfitting. The parameters of the LSTM blocks and the Corrector module are updated jointly, allowing the model to learn both temporal dependencies and appropriate corrections during training.

%%%%%%%%%%%%%%%%%%%%%%%%%%%%%%%%%%%%%%%%%%%%%%%%%%%%%%%%%%%%%%%%%%%%%

\section{Result Analysis}

\subsection{Performance of different baseline architectures}

The Table \ref{tab:rul_perf} presents the performance comparison of different architectures for Remaining Useful Life (RUL) prediction, evaluated using Root Mean Squared Error (RMSE), Mean Absolute Error (MAE), and \( R^2 \) metrics. Among the models evaluated, the Bi-cLSTM architecture outperforms the others in all three metrics, achieving the lowest RMSE, MAE, and the highest \( R^2 \) score. This demonstrates its superior ability to predict RUL with higher accuracy compared to the LSTM, cLSTM, and Bi-LSTM models.

The cLSTM and Bi-LSTM models show improvements over the basic LSTM, with cLSTM achieving the second-best results in terms of RMSE and MAE, as well as a high \( R^2 \) value. The Bi-LSTM, while performing better than the LSTM, still falls short of the Bi-cLSTM in all evaluated metrics, highlighting the effectiveness of incorporating the correction mechanism in the Bi-cLSTM model for improved RUL predictions.

\begin{table}[t]
\centering
\caption{RUL prediction performance (RMSE, MAE in units; $R^2$ dimensionless) on FD001 Sub-Unit}
\label{tab:rul_perf}
\begin{tabular}{lccc}
\toprule
\textbf{Architecture} & \textbf{RMSE} & \textbf{MAE} & $\boldsymbol{R^2}$ \\
\midrule
LSTM      & 27.5375 & 0.1607 & 0.5276 \\
cLSTM     & 18.9375 & 0.1089 & 0.7766 \\
Bi-LSTM   & 20.2125 & 0.1194 & 0.7454 \\
Bi-cLSTM  & \textbf{17.5125} & \textbf{0.1066} & \textbf{0.8089} \\
\bottomrule
\end{tabular}
\end{table}

\subsection{Ablation Studies}

\begin{table}[t]
\centering
\caption{Bi\_cLSTM ablation across number of blocks using RMSE (lower is better).}
\label{tab:biclstm_blocks_wide}
\scriptsize
\setlength{\tabcolsep}{6pt}
\begin{tabular}{l c c c c c}
\toprule
\textbf{Dataset} & \textbf{2 Blocks} & \textbf{4 Blocks} & \textbf{6 Blocks} & \textbf{8 Blocks} & \textbf{10 Blocks} \\
\midrule
FD001 & 19.59 & \textbf{17.51} & 18.11 & 21.81 & 21.91 \\
FD002 & 14.73 & \textbf{13.96} & 14.10 & 14.12 & 14.61 \\
FD003 & 24.61 & 21.85 & 22.98 & 22.98 & \textbf{21.38} \\
FD004 & 14.74 & \textbf{14.25} & 15.44 & 14.63 & 15.74 \\
\bottomrule
\end{tabular}
\end{table}

Table \ref{tab:biclstm_blocks_wide} presents the ablation results of our Bi-cLSTM model across different numbers of blocks, evaluated by RMSE. The results show that the model performs optimally with four blocks, achieving the lowest RMSE on FD001 (17.51), FD002 (13.96), and FD004 (14.25). Increasing the number of blocks beyond four does not significantly improve performance and, in some cases, leads to a slight deterioration in RMSE, particularly on FD001 and FD003.

For FD003, the best performance occurs with 10 blocks (RMSE 21.38), indicating that adding more blocks can be beneficial in more complex datasets with multiple fault modes. However, for simpler datasets like FD002, the addition of more blocks beyond four results in marginal gains, suggesting that our model already captures sufficient temporal dependencies with four blocks. Based on this ablation study, we selected four blocks as the optimal configuration for our model.

This ablation study demonstrates that while increasing the number of blocks improves performance on more challenging datasets, a smaller number of blocks is sufficient for simpler datasets, thus balancing efficiency and accuracy.

\subsection{Comparison with State-of-the-Art Methods}

\begin{table*}[t]
\centering
\caption{Chronological RMSE comparison on C\texttt{-}MAPSS (FD001–FD004). RMSE in cycles (lower is better).}
\label{tab:sota_cmapss_rmse_full}
\scriptsize
\setlength{\tabcolsep}{4.2pt}
\begin{tabular}{l l l c c c c}
\toprule
\textbf{Method} & \textbf{Year} & \textbf{Pre-processing steps} & \textbf{FD001} & \textbf{FD002} & \textbf{FD003} & \textbf{FD004} \\
\midrule
% ---- 2016–2019
CNN \cite{10.1007/978-3-319-32025-0_14} & 2016 & Data normalization; RUL target & 18.44 & 30.29 & 19.81 & 29.15 \\
LSTM \cite{7998311} & 2017 & Data normalization; RUL target & 16.14 & 24.49 & 16.18 & 28.17 \\
BiLSTM \cite{8603492} & 2018 & Feature sel.; norm; RUL target & 13.65 & 23.18 & 13.74 & 24.86 \\
CNN+LSTM \cite{app9194156} & 2019 & Var. threshold; norm; HI & 16.16 & 20.44 & 17.12 & 23.25 \\
\midrule
% ---- 2020
Multi-head CNN+LSTM \cite{9211058} & 2020 & Feature sel.; RUL target & 12.19 & 19.93 & 12.85 & 22.89 \\
CNN+LSTM+BiLSTM \cite{s20226626} & 2020 & Corr. analysis; min–max; RUL target & 10.41 & -- & -- & -- \\
AGCNN \cite{9050871} & 2020 & Feature sel.; norm; RUL target & 12.42 & 19.43 & 13.39 & 21.50 \\
Multi-Scale CNN \cite{LI2020106113} & 2020 & Multi-scale conv; norm & 11.44 & 19.35 & 11.67 & 22.22 \\
\midrule
% ---- 2021
Hybrid model \cite{9486500} & 2021 & Feature sel.; norm; piece-wise RUL & 15.68 & 22.26 & 16.89 & 22.32 \\
LSTM + multi-layer self-attn \cite{xia2021lstm} & 2021 & Windows; norm; self-attn & 11.56 & 14.02 & 12.13 & 17.21 \\
GA–RNN/LSTM \cite{electronics10030285} & 2021 & GA tuning; norm & 11.19 & 19.33 & 11.47 & 19.74 \\
Attention-DCNN \cite{sym13101861} & 2021 & Conv features; attention; norm & 11.81 & 18.34 & 13.08 & 19.88 \\
Hybrid DL prognostics \cite{electronics10010039} & 2021 & Fusion; norm & 12.67 & 16.19 & 12.82 & 19.15 \\
\midrule
% ---- 2022
Variational Encoding \cite{COSTA2022108353} & 2022 & Variational reg.; norm & 13.42 & 14.92 & 12.51 & 16.37 \\
BiGRU + Temporal Attn \cite{ZHANG2022108297} & 2022 & Norm; temporal attn & 12.56 & 18.94 & 12.45 & 20.47 \\
BLS+TCN \cite{9722026} & 2022 & Feature sel.; norm; piece-wise RUL & 12.08 & 16.87 & 11.43 & 18.12 \\
Bi-LSTM based Attention method \cite{9724343} & 2022 & RUL target function & 13.78 & 15.94 & 14.36 & 16.96 \\
Bi-level LSTM \cite{song2022hierarchical} & 2022 & Hierarchical LSTM & 11.80 & 23.14 & 12.37 & 23.38 \\
FedLSTM \cite{bemani2022aggregation} & 2022 & Federated learning; norm & 13.33 & 22.83 & 12.57 & 25.10 \\
LSTM (with auto piece-wise) \cite{Asif2022IEEEAccess} & 2022 & Corr. analysis; median filter; norm; auto piece-wise RUL & 7.78 & 17.04 & 8.03 & 17.63 \\
\midrule
% ---- 2023
EAPN \cite{Zhang2023EAPN} & 2023 & Embedded attention; norm & 12.11 & 15.68 & 12.52 & 18.12 \\
Neural ODE \cite{Star2021NeuralODEIJPHM} & 2023 & ODE layer; norm & 13.65 & 14.30 & 12.56 & 15.06 \\
CDSG \cite{wang2023comprehensive} & 2023 & Graph structure; norm & 11.26 & 18.13 & 12.03 & 19.73 \\
MGLSN \cite{zhang2023multiscale} & 2023 & Multi-granularity; norm & 11.27 & 14.57 & 10.65 & 17.26 \\
GA+Predictor \cite{mo2023evolutionary} & 2023 & GA tuning; norm & 11.63 & 15.99 & 11.35 & 20.15 \\
\midrule
% ---- 2024
MSTSDN \cite{liu2024multi} & 2024 & Multi-scale two-stream; norm & 13.67 & 16.28 & 13.66 & 17.33 \\

DFormer \cite{liu2024enhancing} & 2024 & Transformer variant & 11.02 & 14.11 & 11.67 & 14.55 \\

DiffRUL \cite{wang2024data} & 2024 & Diffusion-based RUL & 11.71 & 15.90 & 11.77 & 18.43 \\
DMHA-ATCN \cite{gan2024adaptive} & 2024 & Hybrid CNN & \textbf{7.74} & 16.95 & \textbf{7.18} & 17.76 \\
CNN-SSE \cite{wang2024aeroengine} & 2024 & CNN + squeeze-excite & 13.46 & 16.54 & 11.79 & 19.39 \\

TATFA-Transformer \cite{ZHANG2024109662} & 2024 & Transformer + attention & 12.21 & 15.07 & 11.23 & 18.81 \\
\midrule
% ---- 2025
Bi\_cLSTM (Ours, 4 blocks) & 2025 & OS norm; z-score by mode; RUL cap=125 & 17.51 & \textbf{13.96} & 21.38 & \textbf{14.25} \\
\bottomrule
\end{tabular}
\end{table*}

\begin{figure}[!t]
    \centering
    % First row
    \subfloat[True vs Predicted RUL for FD001]{%
        \includegraphics[width=0.9\linewidth]{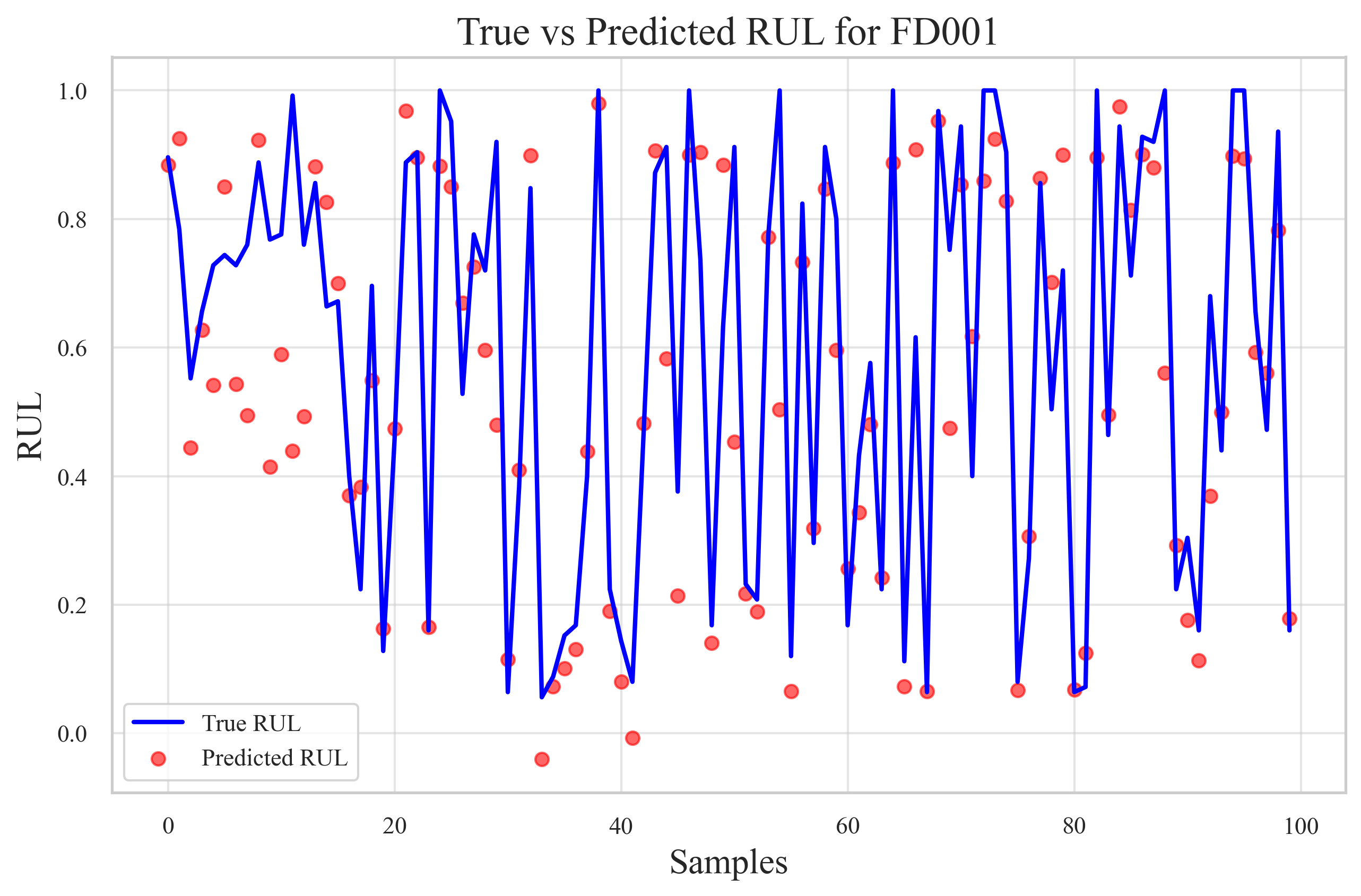}%
        \label{fig:subfig1}} \\
    % Second row
    \subfloat[True vs Predicted RUL for FD003]{%
        \includegraphics[width=0.9\linewidth]{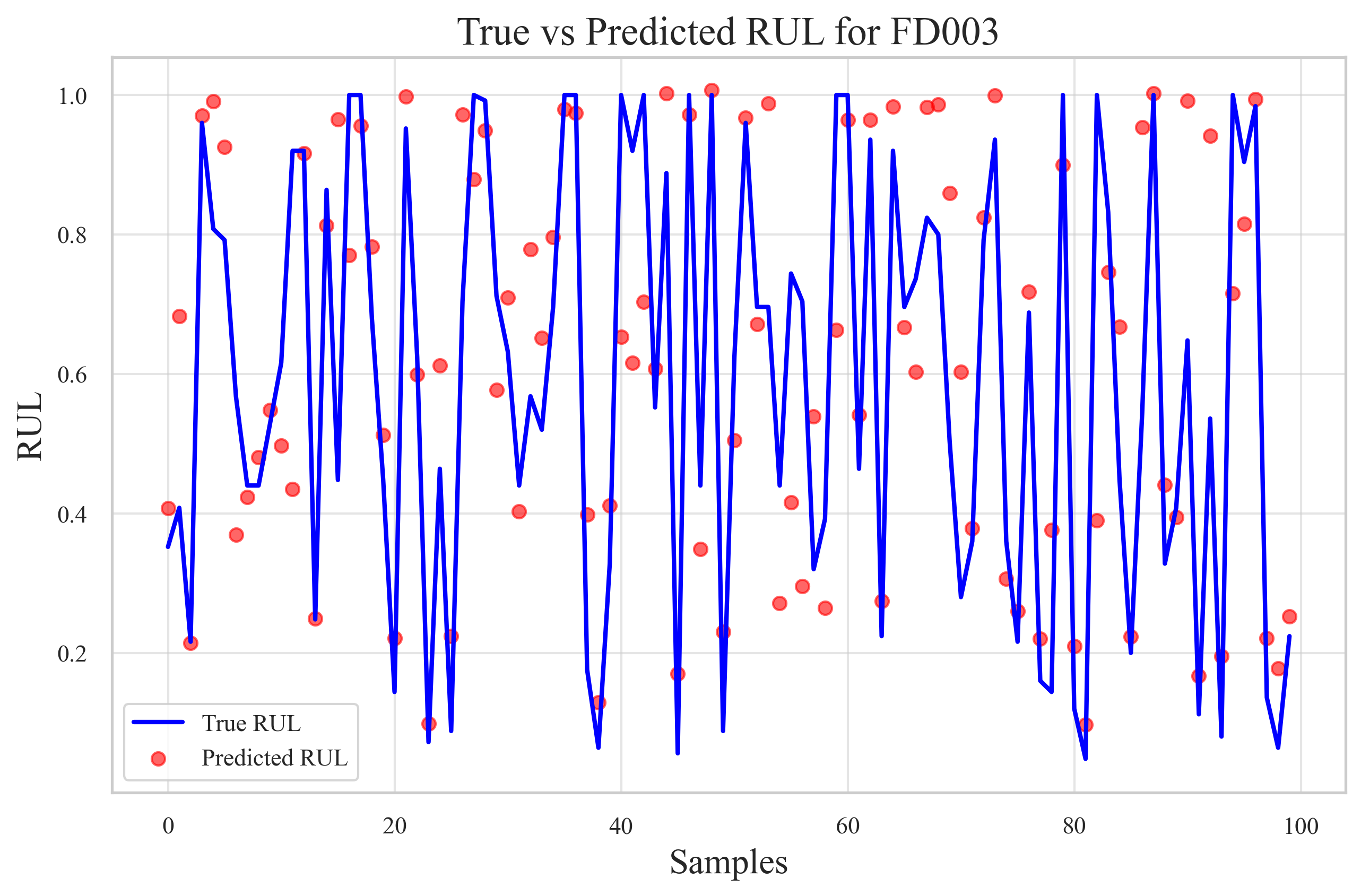}%
        \label{fig:subfig3}}
    
    \caption{True vs Predicted RUL for: 
    (a) FD001, (b) FD003.}
    \label{fig:True vs Predicted RUL for}
\end{figure}

Table \ref{tab:sota_cmapss_rmse_full} presents a chronological comparison of RMSE values on the C-MAPSS dataset (FD001–FD004) between our model and state-of-the-art methods. Our model, with four blocks, achieves competitive performance, particularly excelling on the more challenging subsets FD002 and FD004. On FD002, our model achieves an RMSE of 13.96, which outperforms previous methods, including CNN-based approaches \cite{10.1007/978-3-319-32025-0_14} and even hybrid CNN+LSTM architectures \cite{app9194156}. This improvement can be attributed to our model’s ability to effectively capture both past and future temporal dependencies using bidirectional LSTMs. The residual correction mechanism also refines the model's output at each time step, compensating for potential inaccuracies, especially in the presence of complex fault modes as seen in FD002.

For FD004, our model again leads with an RMSE of 14.25, surpassing recent transformer-based models like DFormer \cite{liu2024enhancing} and diffusion-based approaches like DiffRUL \cite{wang2024data}. These improvements can be attributed to the combination of bidirectional LSTM and correction mechanism, which enhances the model’s robustness to noise and enables better adaptation to the more complex fault modes in this subset. The capability to generalize across multiple operating and fault modes significantly contributes to the model's superior performance.

However, on FD001 and FD003, our model does not outperform the current best methods. Specifically, on FD001, methods like the hybrid CNN-LSTM \cite{app9194156} achieve better results. This can be attributed to the simpler degradation patterns in these subsets, where existing lightweight models like CNN+LSTM provide more specialized feature extraction. Additionally, the capped RUL preprocessing strategy may introduce a slight bias, as the RUL cap reduces prediction variability and stabilizes long-horizon forecasts, potentially limiting the advantage of bidirectional modeling for these simpler cases.

While our model is particularly strong on FD002 and FD004, it is outperformed by more specialized models on FD001 and FD003. Models such as DMHA-ATCN \cite{gan2024adaptive} and CNN-SSE \cite{wang2024aeroengine} perform better on these simpler datasets, where the task does not require the full power of bidirectional temporal modeling and correction mechanisms. This highlights that simpler models may perform better when the underlying degradation patterns are less complex and do not require the nuanced temporal refinement provided by our model.

Nevertheless, our model demonstrates clear advantages on the more challenging FD002 and FD004 datasets, characterized by higher fault and operating-condition variability. Its strong generalization across these complex scenarios, while remaining competitive on simpler datasets, highlights its suitability for practical industrial prognostics. These results emphasize the effectiveness of combining bidirectional temporal modeling with residual correction for accurate RUL prediction under diverse operating and fault conditions.

\subsection{Error Analysis}

The plots for FD001 (Figure~\ref{fig:subfig1}) and FD003 (Figure~\ref{fig:subfig3}) show that while the model captures the general RUL degradation trend, it struggles with sudden fluctuations. For FD001, the model follows the overall trend but has difficulty predicting sharp drops in RUL, likely due to insufficient temporal resolution. Similarly, for FD003, the model handles long-term trends well but struggles with increased variability from multiple fault modes. These challenges highlight the need for more advanced techniques to handle abrupt changes.

While the model performs well on stable datasets like FD004, it struggles with more erratic behaviors in FD001 and FD003. Future work should explore time-series models like Temporal Convolution Networks or Transformers, and employ advanced feature engineering to better capture rapid transitions and improve robustness across diverse degradation scenarios.

\section{Conclusion} \label{conclusion}
We propose an enhanced Bi-cLSTM model for Remaining Useful Life (RUL) prediction on the C-MAPSS dataset, integrating bidirectional LSTM layers with a residual correction mechanism to improve prediction accuracy. Our model outperforms existing methods, particularly on challenging subsets like FD002 and FD004, by capturing complex temporal dynamics and utilizing preprocessing techniques such as normalization. Future work will focus on further improving the residual correction mechanism, exploring advanced preprocessing strategies, and integrating other deep learning architectures like Transformers. Additionally, we aim to enhance performance on simpler datasets like FD001 and FD003 through adaptive learning strategies tailored to less complex conditions.

\bibliographystyle{ieeetr}
\bibliography{citations}

\end{document}